\documentclass{article} 
\PassOptionsToPackage{sort}{natbib}
\usepackage{iclr2021_conference,times}

\usepackage{amsmath,amsfonts,bm}

\def\eqref#1{equation~\ref{#1}}

\def\1{\bm{1}}

\DeclareMathAlphabet{\mathsfit}{\encodingdefault}{\sfdefault}{m}{sl}
\SetMathAlphabet{\mathsfit}{bold}{\encodingdefault}{\sfdefault}{bx}{n}

\usepackage{hyperref}
\usepackage{url}
\usepackage{multirow}
\usepackage{booktabs}
\usepackage{wrapfig}
\usepackage{graphicx}
\usepackage{comment}
\usepackage{caption}
\usepackage{subcaption}
\usepackage{csquotes}
\newcommand*\samethanks[1][\value{footnote}]{\footnotemark[#1]}

\title{Model Selection's Disparate Impact in Real-World Deep Learning Applications}

\author{Jessica Zosa Forde\thanks{Equal contribution.}, A. Feder Cooper\samethanks{}, Kweku Kwegyir-Aggrey, Chris De Sa, Michael Littman\thanks{Author emails: jessica\_forde@brown.edu, afc78@cornell.edu, kweku@brown.edu, cmd353@cornell.edu, michael\_littman@brown.edu}
}

\iclrfinalcopy 
\begin{document}

\maketitle

\begin{abstract}
Algorithmic fairness has emphasized the role of biased data in automated decision outcomes. Recently, there has been a shift in attention to sources of bias that implicate fairness in other stages in the ML pipeline. We contend that one source of such bias, human preferences in model selection, remains under-explored in terms of its role in disparate impact across demographic groups. Using a deep learning model trained on real-world medical imaging data, we verify our claim empirically and argue that choice of metric for model comparison can significantly bias model-selection outcomes. 

\end{abstract}

\section{Introduction}
While ML promised to remove human biases from decision making, the past several years have made it increasingly clear that automation is not a panacea with respect to fairer decision outcomes. The interaction between automation and fairness is exceedingly complex: Applying algorithmic techniques to social problems without nuance can magnify rather than correct for human bias~\citep{abebe2020social}. As a result, various stakeholders, from researchers to policymakers to activists, have criticized algorithmic decision systems and advocated for approaches to mitigate the potential harms they inflict on marginalized groups.

Common to their critiques is the emphasis of data's role in unfair outcomes: Pre-existing bias in datasets results in training biased models; undersampling of marginalized populations results in worse predictive accuracy in comparison to more privileged, represented populations. In other words, regardless of the origin of bias in the data, critics have often considered this bias to be the main factor responsible for automated decision systems' reinforcement of existing inequities~\citep{Buolamwini2018-lt, De_Vries2019-dj, Denton2020-vv}. More recently, some algorithmic fairness stakeholders have begun to shift their focus to alternate sources of unfairness in ML pipelines. They contend that bias is not just in the data, and place increased scrutiny on modelling decisions. In particular, they examine the process of \emph{problem formulation}---how data scientists construct tasks to make them amenable to ML techniques---as a source of bias in ML~\citep{Passi2019-Formulation, cooper2021emergent, hicks2019cistem, Bagdasaryan2019-aq, Hooker2019-ne}. 

In this paper, we highlight another under-examined source of modelling bias---one that exists further down the ML pipeline from problem formulation. \emph{Model selection}, the process by which researchers and data scientists choose which model to deploy after training and validation, presents another opportunity for introducing bias. During model selection for solving a particular task, the model developer compares differences in the performance of several learned models trained under various conditions, such as different optimizers or hyperparameters. This procedure is not a strictly computational; rather, the metrics used to distinguish between models are subject to human interpretation and judgement~\citep{Adebayo2018-zk, Jacobs2019-vq}. Human preferences, often geared toward the particular application domain, ultimately play an important role in choosing the model to deploy. 

This role of human influence on model selection has been explored in relation to reproducibility~\citep{Henderson2018-yx}, ``researcher degrees of freedom" and p-hacking \citep{Gelman2014-be}, and hyperparameter optimization \citep{cooper2021deception}. \textbf{We argue that these choices also have implications for fairness.} 
We provide an intuition for model selection's possible role in disparate impact for different demographic groups (Section \ref{sec:intuition}) and verify this intuition empirically on real-world medical imaging DL tasks (Section \ref{sec:demonstration}). To clarify the relevance and impact of this issue, we close with a call for researchers to place greater emphasis on experiments with real-world datasets and choice of model-selection comparison metrics, as commonly-used benchmark tasks and overall accuracy metrics can obscure model selection's implications for fairness (Section \ref{sec:implications}).

\section{Preliminaries}
\label{sec:prelim}
In ML, we typically fit a chosen model to a training dataset, iteratively improving until convergence. We evaluate the resulting model's performance for some metric, such as overall accuracy or area under the curve (AUC) on a test dataset. To further improve performance, the learning process can be repeated, changing the hyperparameters controlling the optimization process or simply using a different random seed. We then compare the performance of the different models, usually reporting or deploying the one that does the ``best," where ``best" is defined by the chosen performance metric~\citep{hastie2009elements, Tobin2019-bw}. 
The process of repeatedly retraining until a satisfactory model is produced is what \citet{Gelman2014-be} more formally call a ``researcher degree of freedom" in the model-selection process. While researchers may report average performance across a small number of models, they typically report results using a specific set of hyperparameters, which can obscure a model's true performance~\citep{Bouthillier2019-qy, Choi2019-zj, Sivaprasad2020-el, cooper2021deception}. Moreover, when selecting a model to deploy, engineers often select a specific training run, rather than an ensemble of training runs.

While a researcher or engineer may choose between models with similar performance, many other proprieties of models change between training runs. In classification contexts, aside from overall accuracy, \emph{sub-population performance} serves as an important consideration for model selection and can vary even when overall accuracy falls in a small range.
Consider a model trained on CIFAR-10 with 10 classes and equal training data in each class. We can consider each class as a sub-population of the overall dataset. For an overall accuracy of 94\%, there are numerous possibilities for how well a model performs on each of the 10 classes. Each class could exhibit 94\% accuracy or, in a pathological case, 9 of the classes could have 100\% accuracy with the 10th classifying correctly at a paltry 40\% rate. In other words, not all 94\% accuracy models are interchangeable, even for the same model architecture.

\section{The Impact of Model Selection: Intuition} \label{sec:intuition}

While this pathological case does not necessarily occur in practice (Appendix, Figure \ref{fig:cifar10}), smaller, more realistic variability in sub-population performance can still have a significant impact. Consider large-scale, high-impact applications, such as medical imaging. Seemingly minute differences in sub-population performance can multiply out to potentially huge numbers of people with missed diagnoses. Such differences can impact fairness if diagnostic misses disproportionately affect some demographic groups. 

In this vein, we explore a modified example from \citet{Sritisava2019-perceptions}, summarized in Table~\ref{table:modelselection}. The authors surveyed test subjects concerning fair model selection using artificial results for 3 models trained using different algorithms, each yielding different overall accuracy and female/male accuracy rates for skin cancer detection. They asked the subjects to pick the ``fairest" model, and they overwhelmingly selected the model trained by $A_1$. The authors reason that the subjects chose this model because, even though it exhibits the largest disparity been males and females, it has the overall highest accuracy; they posit that, in high-impact domains like medicine, it is preferable to select more accurate models, even those with a potentially high cost to fairness. 

\begin{table}[ht]
\centering
\caption{Algorithms resulting in different model accuracy--fairness trade-offs for skin cancer risk prediction as presented in \cite{Sritisava2019-perceptions}. The authors asked test subjects to choose between the presented models and found that subjects preferred $A_1$ (which has highest overall accuracy).
}
\label{table:modelselection}
\begin{tabular}{c c c c}
\toprule
Algorithm & Overall Accuracy & Female Accuracy & Male Accuracy\\
\midrule
$A_1$ & 94\% & 89\% & 99\% \\
$A_2$ & 91\% & 90\% & 92\% \\
$A_3$ & 86\% & 86\% & 86\% \\
\bottomrule
\end{tabular}
\end{table}
Importantly, this experiment does not take into account underlying sub-population rates of skin cancer, and why overall accuracy might be a misleading metric to consider. To underscore why this oversight is important, consider instead that Table~\ref{table:modelselection} concerns detecting breast cancer instead of skin cancer. When explicitly considering accuracy in terms of sub-population, which algorithm would be fairest? Since overall accuracy only measures label alignment, we do not know the breakdown of false positives and false negatives. In medical domains, false negatives (missing a cancer diagnosis) generally incur a much higher cost than false positives  (mistakenly diagnosing cancer when it is in fact not present). We could therefore argue $A_2$ now has a stronger case: Males get breast cancer, but much less frequently than females~\citep{Giordano2002-al}; depending on these different rates of breast cancer, the extra 1\% in accuracy for females using $A_2$ might make it a better choice than $A_1$ if it corresponds to an overall lower number of false negative (``missed'') cancer diagnoses. We could make a similar argument about how human choices in model selection might change in the context of skin cancer prediction for individuals of different races~\citep{Adamson2018-vg}.\footnote{This example is illustrative only; we recognize that training one model is not the only solution for building an automated decision system. In cases like this one, it is perhaps desirable to train multiple, decoupled models~\citep{dwork2018decoupled}. This regime, however, has its own issues, such as how to assign individuals that span multiple demographic groups to different classifiers.}

The overarching point is that contextual information is highly important for model selection, particularly with regard to which metrics we choose to inform the selection decision. Overall accuracy conceals the decomposition into false positives and false negatives. Including these metrics could bias us to make different choices concerning which model is ``best" in a particular application domain (especially in medical domains, in which the relative costs of false negatives and positives is markedly different). Moreover, sub-population performance variability, where the sub-populations are split on protected attributes, can be a crucial part of that context, which in turn has implications for fairness. In medicine, this variability corresponds to the case where underlying disease rates vary by sub-population. \textbf{We therefore advocate examining sub-population performance variability as an essential component of performing fair model selection.}



\section{A Deep Learning Example from Medical Imaging} \label{sec:demonstration}

\begin{figure}[ht]
\centering
\begin{subfigure}[b]{.5\linewidth}
\includegraphics[width=\linewidth]{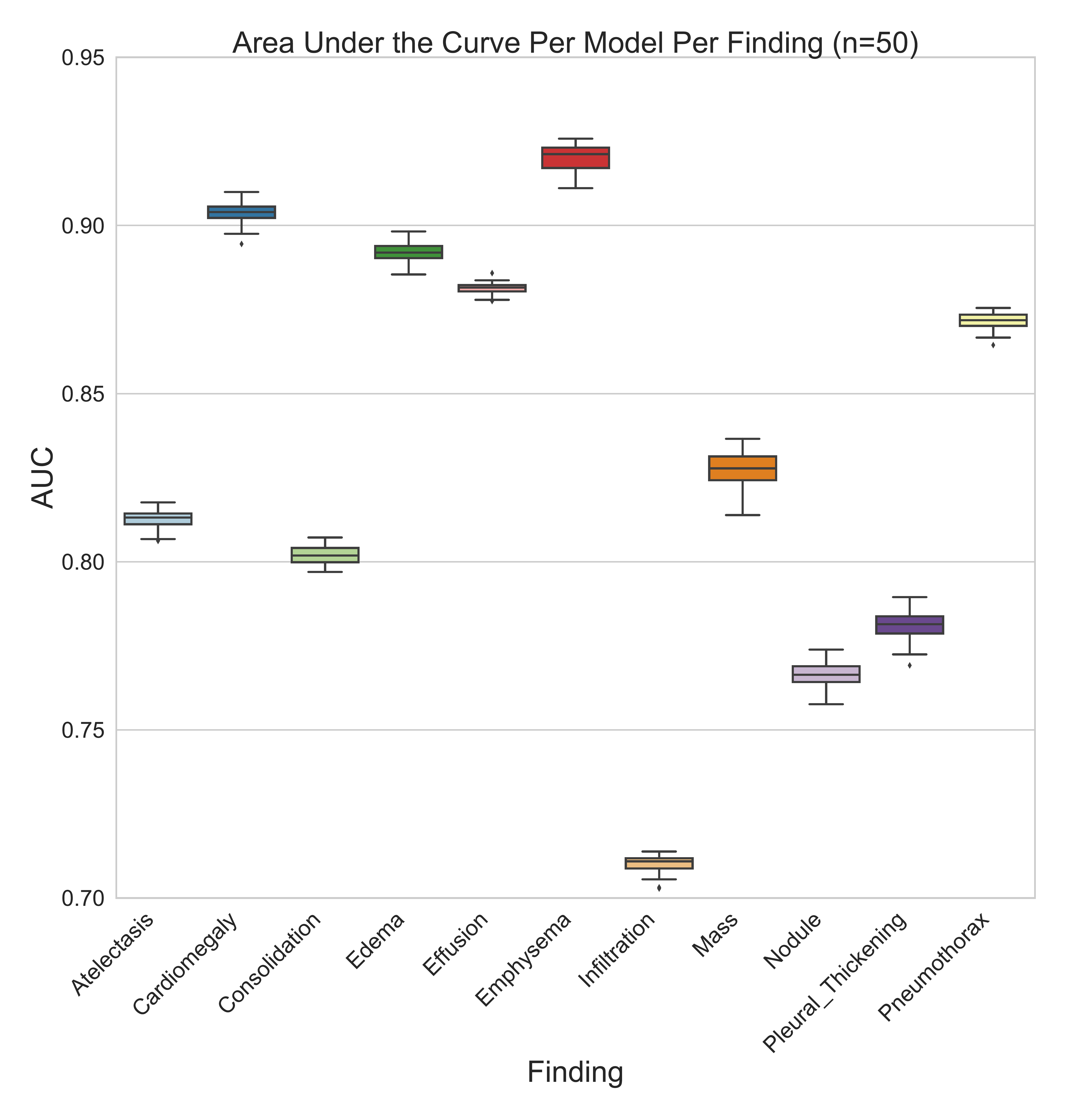}
\caption{Boxplot of Area Under the Curve of CheXNet trained with fixed hyperparameters 50 times, varying random seed with each retraining.  Each boxplot represents 50 models' predictions for the given radiological class (``finding'').}\label{fig:auc}
\end{subfigure}
\hspace{2pt}
\begin{subfigure}[b]{.48\linewidth}
\captionsetup{aboveskip=25pt}
\includegraphics[width=\linewidth]{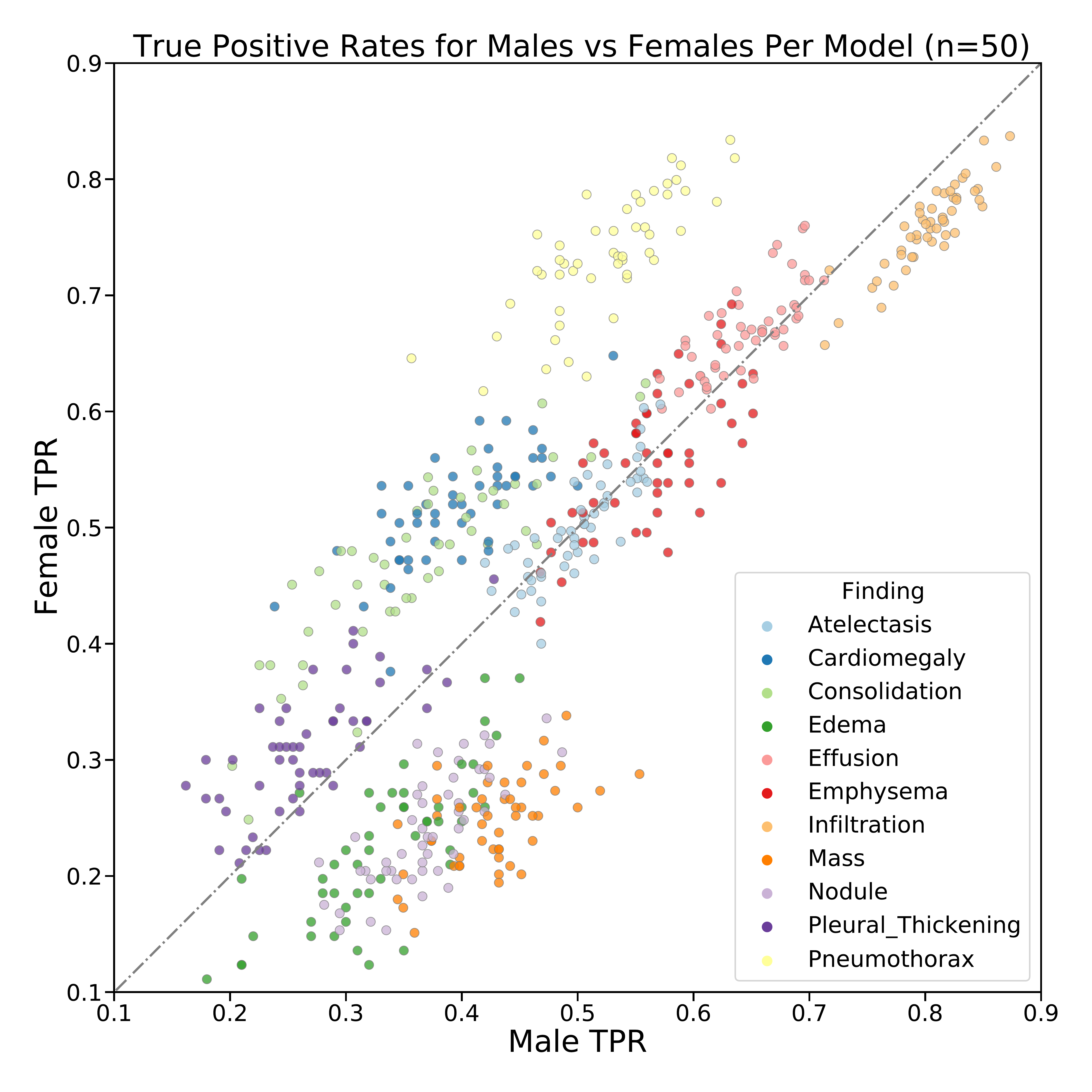}
\caption{True-positive-rate (TPR) disparity among 50 CheXNet models, varying random seed.  Each point represents the TPR for male patients versus female patients for a finding. Points closer to the gray line have lower disparate impact between sexes.\label{fig:tpr}}
\end{subfigure}
\caption{Training CheXNet 50 times with fixed hyperparameters while varying random seed. CheXNet uses a pre-trained initialization, so only batching is affected by random seed.}
\label{fig:chexnet}
\end{figure}

Prior work has demonstrated that medical imaging, like other computer-vision applications, exhibits disparate impact, which can lead to severe disparities in quality of care~\citep{Seyyed-Kalantari2021-ut}. 
We measure the baseline variability of the medical imaging classifier CheXNet, which is used to identify pathologies in chest X-rays~\citep{Rajpurkar2017-iz}. CheXNet fine-tunes DenseNet-121 \citep{Huang2017-hi} with fixed, pretrained ImageNet weights \citep{Russakovsky2015-ib} on ChestX-Ray8~\citep{Wang2017-ow}, a standard, anonymized medical imaging dataset that identifies 14 radiological classes, known as \emph{findings}. Findings are not mutually exclusive, and the incidence of positive test cases can be as low as 0.21\% in the test set. Patient age and sex are included for each image. In this example, we focus on findings with over 50 test examples each for male and female patients. We applied a training and evaluation procedure similar to \citet{Zech2019-wj} and \citet{Seyyed-Kalantari2021-ut}:  We trained 50 CheXNet models with the same optimizer and same hyperparameters, varying the random seed across runs.  Because the pre-trained weights are fixed based on ImageNet pre-training, the resulting models differ only in the order in which the images are input in minibatches to the model. \textbf{Our results suggest that, even among models with the same training procedure, we can observe significant differences in sub-population performance (Figure~\ref{fig:tpr}), while exhibiting little variability in overall performance concerning findings (Figure~\ref{fig:auc})}. 

In particular, we measure overall performance by looking at the AUC for each finding; the largest range in AUC between all models is 0.0227 for the Mass finding. As each model returns a score for each instance for each finding, classification requires a threshold on this score. We select the prediction threshold for each model to maximize the F1 score on training data~\citep{Chinchor1992-rs}. Given each model's specific prediction threshold, we apply each threshold to the test set and measure the true positive rate (TPR) of each model. The choice of TPR over overall accuracy is due to the low prevalence of positive labels for the findings considered (as low as 3.5\% for Edema) and for the importance of correctly classifying positive cases as discussed in Section \ref{sec:intuition}. In optimizing for F1 score, we observe both substantial differences in TPRs for males and females for a given model and finding, but also substantial variability in TPR disparity between models. The biggest TPR disparity between models for a given finding is 0.2892. The largest range of TPR disparities for a given finding across models is 0.2258.

\section{Discussion and Future Work} 
\label{sec:implications}

When we measure sub-population performance levels for CheXNet models with different seeds and similar overall performance, we observe not only high levels of true positive rate disparity, but also find that these disparities can significantly vary between models. 
We therefore contend that such variability suggests additional efforts are necessary when designing for model robustness, especially in relatively under-explored, high-impact domains. Model selection relies on a procedure (either human or automated) for comparisons and choices between algorithmic methods. While we primarily discuss sub-population performance variability within the context of algorithmic fairness, we believe that this line of work has potential for developing more accountable decision systems \citep{Nissenbaum1996accountability}. Specifically, exposing model-selection decisions and the rationales that justify them enables the opportunity to hold those decisions accountable. 

In future work, we intend to formalize this notion of accountability and explore if we can ameliorate the issue of sub-population performance variability that we observe in Figure~\ref{fig:chexnet}. For the latter, we will examine how ensembling---a popular method for improving model performance and reducing prediction variability in medical imaging tasks \citep{Halabi2019-mo, Zech2019-wj}---could level out differences between sub-populations. Moreover, we aim to complement this investigation by studying how AutoML methods, which divest the model-selection process from the whims of human choice, can also prevent subjective preferences from intervening in the hyperparameter optimization process~\citep{Snoek2012-gr}.


\newpage
\section*{Acknowledgments}
We would like to thank the following individuals for feedback on this work: Harry Auster, Charles Isbell, and Michela Paganini. We would also like to thank the Artificial Intelligence Policy \& Practice initiative at Cornell University and the MacArthur Foundation. This work is supported in part by the ONR PERISCOPE MURI award N00014-17-1-2699.

\bibliography{references}
\bibliographystyle{iclr2021_conference}

\newpage
\appendix
\section{Appendix}

\begin{figure}[ht]
\centering
\includegraphics[width=\linewidth]{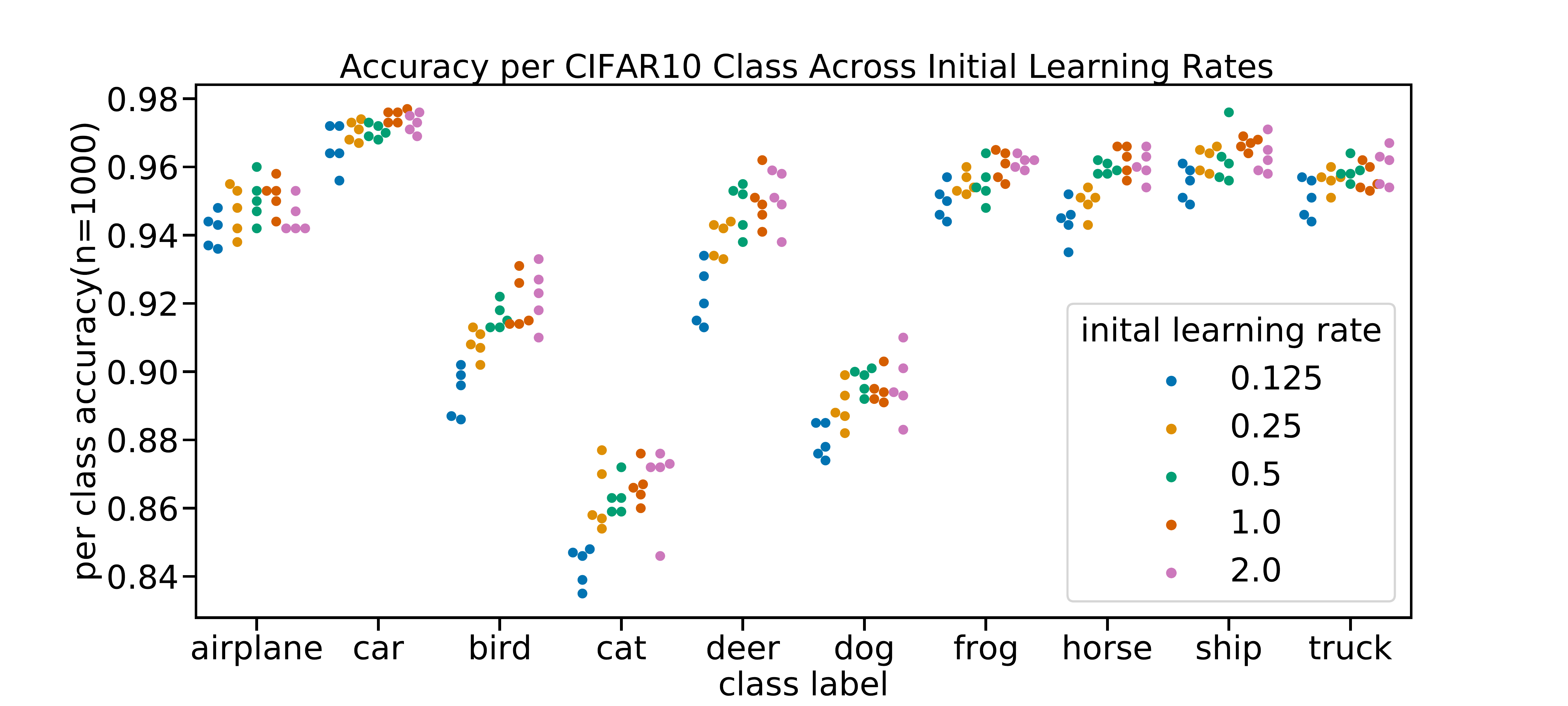}
\caption{Per class accuracy of VGG-16 trained on CIFAR-10.}
\label{fig:cifar10}
\end{figure}

We revisit the toy example of CIFAR-10~\citep{Krizhevsky2009-fx} from Section~\ref{sec:prelim}, where we consider the 10 classes to be different sub-populations. We ran image classification using SGD on VGG-16 following the setup of \citet{Wilson2017-df} (Figure \ref{fig:cifar10}). We did not find similar levels of sub-population performance variability to that of ChexNet. While one can observe declines in sub-population performance when decreasing the initial learning rate, these differences are by percentage points, rather than the tens of percentage points observed in Figure~\ref{fig:tpr}. This outcome is unsurprising; however, given that benchmark datasets like CIFAR-10 are used for designing and tuning off-the-shelf models like VGG-16.

In stark contrast to toy benchmark problems like CIFAR-10, real-world datasets that capture the kinds of problems we actually want to solve with ML are not used to design and tune novel generic classification models. The same amount of collective research effort spent on hyper-optimizing models for CIFAR-10 has not gone to hyper-tuning models like ChexNet. As others have noted, benchmark tasks are not reflective of the novel, difficult problems to which want to apply ML~\citep{De_Vries2019-dj}. In fact, based on our results, we hypothesize that models can be so highly tuned to benchmarks that they conceal problems like sub-population performance variability.

\end{document}